\documentclass[runningheads]{5110}
\usepackage[T1]{fontenc}
\usepackage{graphicx}
\usepackage{microtype}
\usepackage{booktabs} % for professional tables
\usepackage{tikz}

\usepackage{float} 
\usepackage{subcaption}
\usepackage[misc]{ifsym}
\newcommand{\corr}{(\Letter)}
\usepackage{mwe}
\usepackage{caption}
\usepackage{mathtools}
\usepackage{amsfonts}
\usepackage{dsfont}
\usepackage{algorithm}
\usepackage{algorithmic, eucal}
\usepackage{hyperref}
\usepackage{url}
\usepackage{color}

\newcommand{\ie}{\textit{i.e.}}

\makeatletter
\newcommand{\printfnsymbol}[1]{%
  \textsuperscript{\@fnsymbol{#1}}%
}
\makeatother

\begin{document}
\title{Temporal-contextual Event Learning for Pedestrian Crossing Intent Prediction}
% % If the paper title is too long for the running head, you can set
% % an abbreviated paper title here

\titlerunning{Temporal-contextual Event Learning}

\author{Hongbin Liang\inst{1,2} \thanks{Equal contribution} \and
Hezhe Qiao\inst{3} \printfnsymbol{1} \and Wei Huang \inst{4} \and Qizhou Wang\inst{5} \and \\  Mingsheng Shang\inst{1} \and Lin Chen\inst{1} \corr}

\authorrunning{H. Liang and H. Qiao et al.}
% \thanks{Lin Chen is the corresponding author. This work is supported in part by Chongqing Key Project of Technological Innovation and Application (No. CSTB2023TIAD-STX0015). 
% }

\institute{Chongqing Institute of Green and Intelligent Technology, \\ Chinese Academy of Sciences, Chongqing 400714, China \\
\and
Chongqing School, University of Chinese Academy of Sciences, \\ Chongqing 400714, China \\
\email{\{lianghongbin,msshang,chenlin\}@cigit.ac.cn}
\and
Singapore Management University, 178902, Singapore \\
\email{hezheqiao.2022@phdcs.smu.edu.sg}
\and
Beijing University of Posts and Telecommunications, 100876, China \\
\email{huangwei@bupt.edu.cn}
\and
Monash University, 3800, Australia \\
\email{qizhou.wang@monash.edu}
}
\maketitle              % typeset the header of the contribution

\begin{abstract}
Ensuring the safety of vulnerable road users through accurate prediction of pedestrian crossing intention (PCI) plays a crucial role in the context of autonomous and assisted driving. Analyzing the set of observation video frames in ego-view has been widely used in most PCI prediction methods to forecast the cross intent. However, they struggle to capture the critical events related to pedestrian behaviour along the temporal dimension due to the high redundancy of the video frames, which results in the sub-optimal performance of PCI prediction. Our research addresses the challenge by introducing a novel approach called \underline{T}emporal-\underline{c}ontextual Event \underline{L}earning (TCL). 
The TCL is composed of the Temporal Merging Module (TMM), which aims to manage the redundancy by clustering the observed video frames into multiple key temporal events. Then, the Contextual Attention Block (CAB) is employed to adaptively aggregate multiple event features along with visual and non-visual data. By synthesizing the temporal feature extraction and contextual attention on the key information across the critical events, TCL can learn expressive representation for the PCI prediction.  
Extensive experiments are carried out on three widely adopted datasets, including PIE, JAAD-beh, and JAAD-all.
The results show that TCL substantially surpasses the state-of-the-art methods. 
Our code can be accessed at \url{https://github.com/dadaguailhb/TCL}.

\end{abstract}
\section{Introduction}
In the rapidly advancing field of autonomous and assisted driving \cite{khan2022level,lorenzo2021capformer}, the paramount concern is human safety. Facilitating effective interactions between vehicles and vulnerable road users is essential. Within this context, accurately identifying pedestrian crossing intention (PCI) is especially vital. In reality, pedestrian behavior is affected by various factors~\cite{yang2021crossing,ham2022mcip}, such as traffic signs, vehicle speeds, and the conduct of other traffic participants, making the accurate prediction of pedestrian behavior a challenging task. 
Over the past few years, a variety of RNN-based methods for PCI have been introduced, which analyze both visual and non-visual input data, achieving improved performance in PCI prediction \cite{kotseruba2020they,kotseruba2021benchmark,yang2022predicting,ham2022mcip}. 
% \hz{Here give a brief introduction of previous nn-based methods.  }
% On the other hand, a sequence of temporal information enables us to capture dynamic scene details like pedestrian trajectories, thereby improving the ability to predict PCI. 
% \hz{Highlight the motivation in the introduction,  1. ViT struggles to capture high redundancy 2.  xxxx please rewrite this part, clarify the second limitation.}
% Considering xxx (\hz{the limitation of RNN-based methods}),  some works leverage the transform to tackle xx . 
Given that RNNs have a limited capacity to retain global information, often leading to information loss when processing temporal data, some works utilize transformer-based architectures to more effectively capture long-range dependencies and enhance overall performance \cite{lorenzo2021capformer,zhou2023pit,zhang2023trep}. 
Although RNN-based and transformer-based PCI prediction methods have achieved a certain level of success,
they typically extract features from all the observed video frames to forecast a pedestrian’s intention to cross the road. \cite{dong2024pedestrian,lv2024pedestrian}. This approach leads to high redundancy since the adjacent frames often exhibit significant similarity. The high redundancy of temporal sequence data makes the model struggle to capture essential information, resulting in sub-optimal performance.

To deal with the challenges above,  we propose a new network, called temporal-contextual event learning (TCL), illustrated in Fig. \ref{fig:1}, which differs from the prior approaches like RNNs and transformer-based approaches that analyze the whole observed video frames in PCI prediction.
% As illustrated in Fig. \ref{fig:1}, different from the previous method that analyzes the whole observed sequence,  
We first propose the Temporal Merging Module (TMM), which employs event clustering to categorize the observed video frames into multiple key events according to the behavior changes of pedestrians, such as standing, walking, or turning around.  Furthermore, we introduce a contextual attention block (CAB) to explore the relation of multiple critical events, aggregating the critical features at both the event level and the data level. 
%Specifically, the critical event features are learned by adaptively fusing features across events, incorporating both visual and non-visual data.  
% The attention mechanism is employed to weigh each event, thereby forcing the TCL to further extract the critical temporal clues to improve PCI. 
Finally, by leveraging the TMM and CAB, the critical features can be effectively captured in the expressive representation learning of pedestrians, enhancing the accuracy of PCI prediction.
% to address the issue of attention dispersion caused by the high redundancy of temporal sequence data,   % \hz{TODO. Give a description of Fig.1 , TCL focus on the critical event feature extraction making xxx thereby benefits the cross intent detection}
Our primary contributions in this work are:
\begin{figure}[t]
	\centering
	\includegraphics[width=3in,height=1.9in]{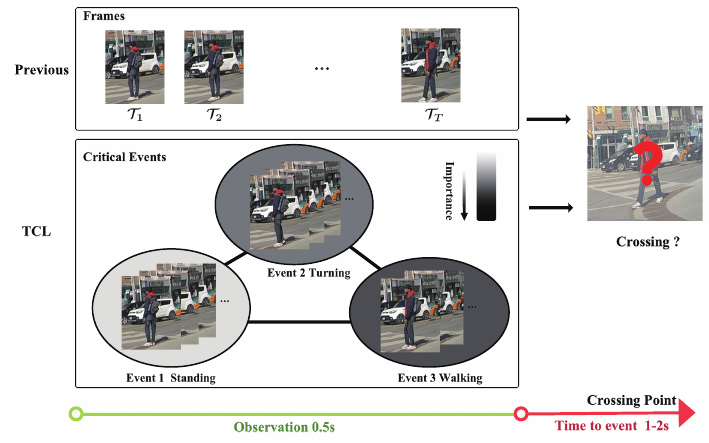}
	\caption{TCL focuses on the critical events identification and adaptively critical features extraction across the events for PCI prediction, contrasting the previous methods that analyze the whole observed video frames. }\label{fig:1}
\end{figure}
\begin{itemize}
\item  We first introduce the temporal merging module to effectively identify the critical information by clustering the temporal frames into multiple critical events.
% aggregating the temporal features based on the segmented events.
% manage redundancy in sequential video data,

\item  We then develop a contextual attention block to adaptively aggregate the critical features from the key events along with visual and non-visual data.

% integrating both visual and non-visual event features derived from corresponding data sources, adaptively fusing limited key events from visual and non-visual inputs to benefit TCL by focusing on critical dynamic changes, thereby improving the PCI. 
% \hz{How the fusion module benefits the PCI, it is necessary to indicate the reason. }

\item  Conducting thorough studies on three prominent datasets, we illustrate that the proposed TCL significantly surpasses existing state-of-the-art approaches.
\end{itemize}
% \subsection{A Subsection Sample}
% Please note that the first paragraph of a section or subsection is
% not indented. The first paragraph that follows a table, figure,
% equation etc. does not need an indent, either.

% Subsequent paragraphs, however, are indented.

% \subsubsection{Sample Heading (Third Level)} Only two levels of
% headings should be numbered. Lower level headings remain unnumbered;
% they are formatted as run-in headings.

\section{Related Work}
% \hz{In this section, categorize current PCI methods and describe them respectively.  Some content in the intro could be moved here.  At the end of each type of methods, indication the difference between TCL and these methods.}

\subsection{RNN-based Cross Intent Detection }
Due to the strong capability of modeling temporal dependencies and relationships between elements, RNN has been widely used for analyzing the video frame sequence \cite{rasouli2019pie,lorenzo2020rnn}.
% Early models, which only analyzed static images or vehicle speed \cite{rasouli2019pie}. Rasouli et al. \cite{rasouli2017they} analyzed single-frame images using 2D CNNs to capture spatial information. This approach was later enhanced by 
Bhattacharyya et al. \cite{bhattacharyya2018long} introduce RNNs to sequence 2D features, thus integrating temporal data into the analysis. Kotseruba et al. \cite{kotseruba2021benchmark} further employ the attention mechanism \cite{luong2015effective} to refine these models, enabling a focused analysis of key temporal and spatial details.
Over time, the complexity and variety of input features used in RNNs have significantly evolved to enhance the performance of PCI prediction \cite{kotseruba2021benchmark,osman2022early,yang2022predicting,yao2021coupling}. 
More non-visual features have been incorporated into the PCI prediction model, such as pedestrian bounding boxes with pose keypoints \cite{kotseruba2021benchmark,fang2018pedestrian}, traffic objects \cite{yao2021coupling,kotseruba2020they}, and contextual segments \cite{yang2022predicting}. Ham et al. \cite{ham2023cipf} propose CIPF, which utilizes eight different input modalities. Yang et al. \cite{10418196} utilize graph convolutional neural networks (GCN) to analyze pedestrian poses and deploy RNN to examine temporal sequences. Additionally, advancements in feature extraction technologies, such as C3D \cite{kotseruba2021benchmark}, have enabled direct joint analysis of spatial and temporal aspects using video data. 

\subsection{Transformer-based Cross Intent Detection }
RNNs or C3D,  while effective in certain tasks, have a limited ability to retain global information\cite{zhao2024review}, typically remembering only the content from recent sequences, which leads to information loss when processing temporal data. 
% This makes it difficult to manage and synthesize comprehensive temporal information effectively. 
In contrast, vision transformer (ViT) architecture networks are capable of building long-range dependencies between images, offering a clear overview of the global context for video recognition. Recently, some ViT-based methods have been introduced to improve the PCI prediction and obtained convincing results \cite{lorenzo2021capformer,lorenzo2021intformer}. Zhang et al. \cite{zhang2023trep} introduce a transformer-based evidential prediction method aimed at uncertainty-aware estimation of pedestrian intentions. Zhou et al. \cite{zhou2023pit} employ stacked transformer layers where each layer corresponds to a time step.  
Nonetheless, although transformer-based models provide a robust overview of the global context for video recognition, the high similarity between video frames can lead to dispersed attention when the attention mechanism is applied to all frames. 

To address this limitation, we built upon video-ViT, using TMM to segment the frame sequence into key events. Then, we applied the contextual attention mechanism to aggregate the critical features adaptively. The proposed TCL can significantly focus on the important information reflected in the data as noticeable dynamic changes.

\section{Methodology}

\subsection{Problem Definition}\label{AA} 
% \hz{In this part, apart from the introduction of input data, pls 
% add the introduction of this task. what does PCI do, some formula or letter can be used to describe this.}
% This research focuses on leveraging the neural network to exploit given video data.
% This cross-intent detection is formulated as,  
Considering a series of observed video frames$\{\mathcal{T}_1, \mathcal{T}_2,..\mathcal{T}_T\}$ where $n$ is the number of observed frames. 
The goal of cross-intent detection is to design a model  $f: \{\mathcal{T}_1, \mathcal{T}_2,..\mathcal{T}_T\} \rightarrow \mathbb{R}$ to predict the probability of the pedestrian's action. 

Due to the significant impact of dynamic environments on pedestrian motion~\cite{karasev2016intent}, numerous explicit features are widely considered in research. 
These features can be broadly categorized into visual data and non-visual data.

% A recent study has demonstrated that dynamic environments have a significant impact on pedestrian motion. 
% Some explicit features have been widely used in the proposed methods such as the pose keypoint, the local context, and the bounding box of the pedestrian. These features can roughly divided into two types including Visual and Non-visual data.  
% Regarding pedestrians in the scene, given \(T\) as the observation duration, the factors influencing the pedestrian’s intention of crossing are as follows:

1) \textbf{Visual data.} The visual data, denoted as \(\{I_1, I_2, \ldots, I_T\}\), mainly consists of the $i$-th pedestrian's local context image, which refers to a square zone delineated by an expanded pedestrian bounding box ("bbox"), including both the pedestrian and contextual elements such as ground, curb, crosswalks, etc.

2) \textbf{Non-visual data.} The Non-visual data, denoted as \(\{I^{\prime}_1, I^{\prime}_2, \ldots, I^{\prime}_T\}\), include the bounding box, pose keypoints of $i$-th pedestrian, and the traffic objects in the scene which are described as the following.

Bounding Boxes:  Represented by \([x_1, y_1, x_2, y_2]\), the position coordinates of a target pedestrian indicate that \((x_1, y_1)\) are the top-left and \((x_2, y_2)\) are the bottom-right coordinates of the pedestrian’s bounding box.

Pose keypoints: The movements of target pedestrians by extracting pedestrian pose keypoints.  Following the previous studies \cite{fang2018pedestrian,ham2022mcip,ham2023cipf}, we apply a pre-trained OpenPose model \cite{cao2017realtime} to obtain body keypoints denoted as $P = \{p^1_i, p^2_i, \ldots, p^T_i\}$ for pedestrian. Each keypoint $P$ consists of a 36-dimensional vector, which includes the 2D coordinates for 18 different pose joints.

Traffic objects: The traffic object refers to the features which can capture specific aspects of the driving environment, including the Traffic Neighbor Feature (\(f_{tn}\)), the Traffic Light Feature (\(f_{tl}\)), the Traffic Sign Feature (\(f_{ts}\)), the Crosswalk Feature (\(f_{c}\)), the Station Feature (\(f_{s}\)) and the Ego Motion Feature (\(f_{e}\)).

\begin{figure*}[t]
	\centering
	\includegraphics[width=1\textwidth]{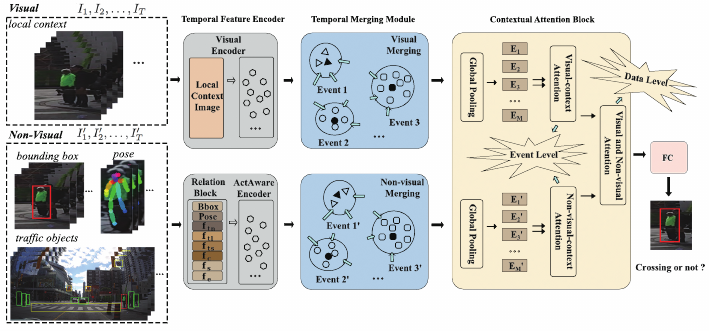}
	\caption{Overview of TCL. It consists of the Temporal Feature Encoder for visual feature extraction, the Temporal Merging Module for frame clustering, and the Contextual Attention Block for the temporal event feature fusion at both event level and data level.
}\label{fig:framework} 
\end{figure*}

% These spatial features combined with the 
\subsection{Network Structure}
% The overall architecture of the proposed model is illustrated in Fig. \ref{fig:framework}. It consists of the temporal feature encoder implemented by the Visual Encoder and the ActAware encoder which aim to extract the visual and non-visual features respectively. 
Fig. ~\ref{fig:framework} illustrates the overall architecture of the proposed model, which comprises three essential components: the feature encoder, the temporal merging module, and the contextual attention block.
The feature encoder consists of the temporal feature encoder implemented by the Visual Encoder and the ActAware Encoder, which aim to extract the visual and non-visual features, respectively.
The temporal merging module clusters the frame into multiple events to exploit the relation of frames, making the model focus on the key dynamic changes in critical events. The contextual attention block employs the attention mechanism to fuse the contextual information and improve the temporal feature extraction at both the event level and the data level.

% To enhance the aggregation of temporal features, the TMM and the FAB are employed to efficiently fuse the temporal clues based on the Temporal Feature Encoder. 

\subsection{Temporal Feature Encoder}
% \hz{Here add one sentence to give a high-level description of temporal feature encoder.Done}
The temporal feature encoder includes both the Visual Encoder and Non-visual Encoder, namely the Actaware Encoder, which is combined with the Relation Block. These encoders help the model learn the representation from the visual and non-visual data, respectively. Then, the visual and non-visual features are fed into the temporal merging modules.
%The pedestrian context images in the visual component, pose data, and various traffic objects in the non-visual set were utilized to enhance prediction performance.

% Given a sequence of images around each target pedestrian with a length of \(T\),
\textbf{Visual Encoder.}  A pre-trained video-ViT network was used as the Visual Encoder to extract the visual feature \cite{tong2022videomae}. Following previous studies, \cite{dosovitskiy2020image,he2022masked,tong2022videomae},  the local context images are resized into a resolution \(224 \times 224\). The ViT-B/16-based model, pre-trained on Kinetics-400, is chosen as the backbone due to its high efficiency and accuracy.
% Therefore, we obtain the extracted feature tokens \(1568 \times 768\), where 1568 (\(8 \times 196\)) represents the number of tokens and 768 is the dimension of each token. Then, we average the token features belonging to the same image to get frame level features \(8 \times 768\).

% In the context of binary classification tasks within neural networks, the binary cross-entropy loss function is a commonly used metric to measure the performance of the model's predictions. This loss function is particularly suitable when the model outputs probabilities that indicate the likelihood of input samples belonging to a specific class. 

\textbf{Non-Visual Encoder.} The Non-Visual data, including the bounding box, pose key points, and traffic objects, are the input of the Non-Visual Encoder consisting of the Relation Block and ActAware Encoder.

1) Relation Block. With the Relation Block, the bounding box and each traffic object feature were initially scaled to a uniform size using a corresponding Fully Connected (FC) layer. Subsequently, these features were concatenated together with pose keypoints, resulting in the new feature $f_r \in \mathbb{R}^{T \times (d_1 \times (n + 1) + d_2)}$, where \(T\) is the observed length, \(d_1\) is the embedding dimension of each traffic object, \( n \) represents the number of categories for traffic object features and \(d_2\) represents the dimension of pose keypoints. 

% The Action Aware branch processes all non-visual data relevant to pedestrian intent except pedestrian images, including the target pedestrian's bounding box, pose keypoints, and other traffic objects. 

% Specifically, \( n \) equals 6 in the PIE dataset and 5 in the JAAD dataset. 

2) ActAware Encoder. The ActAware Encoder is implemented based on transformer architecture, which was originally introduced for the tasks in natural language processing \cite{vaswani2017attention}. It receives the relation block’s output as input and utilizes a transformer-based network with a specification similar to ViT-small \cite{tong2022videomae} to learn the representation of non-visual tokens relevant to pedestrian intent. Finally, it yields the new feature vector of size $f_a \in \mathbb{R}^{T \times d}$, where $T$ represents the observed length and $d$ represents the feature dimension at each time step.

% where \(p\) is a 36-dimensional vector representing the 2D coordinates of 18 pose joints.

\subsection{Temporal Merging Module}

% \textbf{Desribe the limitation of General ViT, especially strength the point they fail to reflect the temporal information}
After feature extraction by the Temporal Feature Encoder, we obtained visual and non-visual features with a length of T. Subsequently, these T features will be clustered utilizing the TMM.

Although video-ViT is a powerful global attention mechanism for the long series of action predictions, it struggles to prioritize the critical information across the observed frames. The high redundancy significantly hinders the PCI. Therefore, we propose a temporal merging module (TMM) to deal with this concern and make the model focus on the key events. It first categorizes the observed frame into different events by employing a density peaks clustering algorithm \cite{du2016study} based on k-nearest neighbors.
% Before feeding the temporal information into the FAB, we segment it into a limited number of events, deriving from \cite{jin2023chat}, to merge certain meaningless and redundant features, which can concentrate on key information to obtain more compact and high-level features. 
% We employ a density peaks clustering algorithm \cite{du2016study} based on k-nearest neighbors, to cluster the event frames.
Clustering is performed on both visual and non-visual data. Here we take the clustering on visual features as an example, starting with the $T$ time-step features \( \mathcal{I} = \{I_t\}_{t=1}^{T} \) derived from the Visual Encoder, we initially compute the local density \( \rho_t \) for each \( I_t \) based on its K-nearest neighbors, which is formulated as:

\begin{equation}
\rho_t=\exp\big(-\frac{1}{K}\sum_{I_k\in\mathrm{KNN}(I_t,\mathcal{I})}\|I_k-I_t\|^2\big), 
\end{equation}
where \(\mathrm{KNN}(I_t,\mathcal{I})\) represents the K-nearest neighbors(excluding itself) of \(I_t\) in \(\mathcal{I}\). We also need to calculate \( \delta_t \) of \( I_t \) representing the shortest distance from point \( I_t \) to any other point that has higher density, which are defined as follows:

\begin{equation}
\delta_t=\begin{cases}\min\limits_{m:\rho_m>\rho_t}\lVert I_m-I_t\rVert^2,&\text{if} \exists m \text{ s.t. } \rho_m>\rho_t.\\\quad\max\limits_m \lVert I_m-I_t\rVert^2,&\text{otherwise.}\end{cases}
\end{equation}

The center of each cluster is selected from points with relatively high \( \rho_t \) and high \( \delta_t \). Then we allocate other points to their closest cluster center and obtain the clusters  $\mathcal{E} = \left\{ {{E_1},{E_2}, \ldots ,{E_M}} \right\}$, representing the multiple events, where $M$ denotes the number of events. 
% and use the average pooling within each cluster to represent the cluster.
% where the result is denoted as \(E\).  $\mathbf{E} = \left\{ {{E_1},{E_2}, \ldots ,{E_N}} \right\}$ denotes the visual events,
Ultimately, the non-visual events $ \mathcal{E^{\prime}} = \left\{ {{E^{\prime}_1},{E^{\prime}_2}, \ldots ,{E^{\prime}_M}} \right\}$ can be obtained using the same principles.

\subsection{Contextual Attention Block (CAB)}
% \hz{Add the reason why we need the CAB according to the motivation}
After obtaining the visual and non-visual event features, we need to fuse the features between events and integrate the features from the visual and non-visual branches using the CAB.

In order to make the model focus on the critical feature after temporal feature extraction, we further employ the CAB to adaptively aggregate across event features, which selectively emphasizes specific aspects of features. Contextual attention is applied on both the event level and the data level. Here, we still take the visual feature $\mathcal{E} = \left\{ {{E_1},{E_2}, \ldots ,{E_M}} \right\}$ as an example, the attention weights for each event is defined  as follows:

\begin{equation}
\alpha_i=\frac{\exp(score(E_i, E_j))}{\sum\limits_{j=1}^{M}\exp(score(E_j, E_{i}))}, 
\end{equation}

where \( score(E_i, E_j) = {E_i}^\top{\bf{W}}_{c}^1E_j \) represents similarity between evener $E_i$ and $E_j$, \({\bf{W}}_{c}^1\) is a trainable weight matrix.  Combined with the weighted sum of features from all preceding time steps, denoted as \(E_p\). It is noted that the same principles are applied to non-visual features. The aggregated features of all events  $E_p$ are the new features of the events.

\begin{equation}
F=tanh({\bf{W}}_{p}^1[E_p;E_M]), E_{p}=\sum\limits_{m}\alpha_{m}E_{m},
\end{equation}
where \({\bf{W}}_{p}^1\) is the trainable parameters mapping the representation into a new space.

Apart from utilizing the attention mechanism at the data level by adaptively fusing the features from the visual and non-visual branches. Then, we obtain the overall visual and non-visual event features, denoted as \( F \) and \( F^{\prime} \), respectively. Correspondingly, the attention weight of both features is calculated as follows:
 
\begin{equation}
\lambda_1=\frac{\exp(score(F^{\prime},F))}{\exp(score(F^{\prime},F)) + \exp(score(F^{\prime},F^{\prime}))}, 
\end{equation}

\begin{equation}
\lambda_2=\frac{\exp(score(F^{\prime},F^{\prime}))}{\exp(score(F^{\prime},F)) + \exp(score(F^{\prime},F^{\prime}))}, 
\end{equation}

where \( score(F^{\prime},F) = {F^{\prime}}^\top{\bf{W}}_{c}^2F \) with the trainable matrix \( {\bf{W}}_{c}^2 \). The ultimate integration of visual and non-visual features $F_p$ is computed as:

\begin{equation}
F_p=\lambda_{1}F + \lambda_{2}F^{\prime},
\end{equation}
where \({\bf{W}}_{p}^2\) is a trainable matrix. The aggregated features are subsequently processed by a fully connected (FC) layer to predict the intention to cross.
\begin{equation}
    P=Sigmoid({\bf{W}}_{p}^2[F_p;F^{\prime}]),
\end{equation}
where $P = \{p_1, p_2, ..., p_N\}$ is the output of the model representing predicted probabilities of crossing.

\subsection{Loss Function}
we employ binary cross-entropy loss in our PCI prediction task to quantify the discrepancy between the actual labels and the predicted probabilities.
\begin{equation}
L = -\frac{1}{N} \sum_{i=1}^N [y_i \log(p_i) + (1 - y_i) \log(1 - p_i)],
\end{equation}
where \( N \) denotes the sample size, \( y_i \) represents the ground truth label of the \( i \)-th sample, and \( p_i \) represents the predicted likelihood that the \( i \)-th sample belongs to the positive class.

\section{Experiments}
In this part, the proposed TCL method for predicting pedestrian crossing intent is evaluated for its effectiveness by conducting experiments that compare it against leading state-of-the-art methods.

\subsection{Datasets and Experimental settings}
\noindent \textbf{Datasets.} Following the previous studies \cite{yang2022predicting,yao2021coupling}, 
evaluation of the proposed model and competing methods is conducted on the two large public naturalistic traffic video datasets, Pedestrian Intent Estimation (PIE) \cite{rasouli2019pie} and Joint Attention in Autonomous Driving (JAAD) \cite{rasouli2017they}.
% \textbf{Give a short description of each dataset, similar to  \cite{yang2022predicting, yao2021coupling} }
The PIE dataset, comprising six hours of driving footage captured by an onboard camera, includes 1,842 pedestrians annotated with 2-D bounding boxes and behavioural tags at 30Hz. Ego-vehicle velocity was obtained using gyroscope measurements collected by the camera. The JAAD dataset comprises two subsets: JAAD Behavioral Data (JAAD-beh) and JAAD All Data (JAAD-all). JAAD-beh includes 495 samples of pedestrians crossing and 191 samples of pedestrians intending to cross., while JAAD-all encompasses an extra 2,100 samples of pedestrians performing non-crossing actions. Please refer to Tab. \ref{table:VI} for further details on dataset statistics.
% More statistical results of the datasets are shown in Tab. \ref{table:VI}
% \hz{May add a dataset table like  paper MCIP: Multi-Stream Network for Pedestrian Crossing Intention Prediction. Add more description of dataset.Done}

\begin{table}[t]
    \centering
    \setlength{\abovecaptionskip}{5pt} % 调整标题与表格的间距
    \caption{Comparison of properties between PIE, and JAAD Datasets}\label{table:VI}
    \renewcommand{\arraystretch}{1.3}
    \scriptsize
    \tabcolsep=4pt
    \begin{tabular}{l|l|l}
    \hline
     & PIE \cite{rasouli2019pie} & JAAD \cite{rasouli2017they}  \\
    \hline
     Number of annotated frames& 293K & 75K  \\
     Number of pedestrians & 1.8K & 2.8K  \\
     Number of pedestrians with behavior annotation & 1.8K & 686 \\
     Number of pedestrians bboxes & 740K & 391K \\
     Average pedestrian track length & 401 & 140  \\
     Ego-vehicle sensor information & yes & no  \\
    \hline
    \end{tabular}
\end{table}

\noindent \textbf{Competing Methods.} To verify the performance of TCL, we compared it against state-of-the-art PCI prediction models. 
MultiRNN \cite{bhattacharyya2018long} utilizes distinct RNN streams to independently process each feature type. SingleRNN \cite{kotseruba2020they}  identifies key visual factors, including crosswalk location and pedestrian orientation, and integrates them into a single RNN model. SF-GRU \cite{rasouli2020pedestrian} utilizes stacked GRUs to receive the hidden states from the GRUs in the lower layers as their input. PCPA \cite{kotseruba2021benchmark} utilizes a C3D network to capture local context images and employs GRUs to analyze critical elements like poses. CAPformer \cite{lorenzo2021capformer} employs two distinct transformer-based encoders to extract features from video sequences and other modalities, respectively. The Coupling Intent and Action (Coupling)\cite{yao2021coupling} model utilizes stacked GRUs to handle temporal sequences, incorporating both current actions and predictive outcomes to enhance pedestrian crossing prediction. The Predicting model \cite{yang2022predicting} extracts features from pedestrian context images and utilizes Hybrid fusion to integrate features from the two branches effectively. MCIP \cite{ham2022mcip} categorizes five inputs, including a segmentation map, into visual and non-visual modules, utilizing an attention mechanism to discern crossing intentions. PIT \cite{zhou2023pit} considered local pedestrian, the global environment, and the movement of ego-vehicle simultaneously. Hybrid-Group \cite{dong2024pedestrian} introduces a novel hybrid fusion method and incorporates two additional dynamic attributes. TREP \cite{zhang2023trep} proposed an innovative transformer-based model to capture temporal correlations and address uncertainty. PFRN \cite{lv2024pedestrian} develops a novel RNN architecture that combines spatial and temporal feature fusion.

\noindent \textbf{Evaluation Metric.} PCI is essentially a binary classification problem, determining whether a pedestrian will cross or not based on the observed frames.  We employ several metrics for evaluation, including accuracy, Area Under the Curve (AUC), F1 score, precision, and recall, which are widely recognized and commonly employed in related research \cite{rasouli2022multi,yang2022predicting,kotseruba2021benchmark,yao2021coupling}.

\noindent \textbf{Implementation Detail.}
TCL is implemented in Pytorch 1.11.1 with Python 3.8. In TCL
% The proposed model was developed and evaluated using the Pytorch framework.
we utilized a resized \(224 \times 224\) resolution as the input image size for the video-ViT backbone, which was initially pre-trained on Kinetics-400 \cite{tong2022videomae}. In the Relation Block, the embedding dimension of each traffic object is 32. Besides, we employ the ActAware Encoder, a transformer-based network, to process non-visual information. This network is designed with an embedding dimension of 384, 6 attention heads, and 12 layers in depth. 

Following the previous studies, for each target pedestrian, we sample observation data, ensuring that the final observed frame is captured within a 1 to 2-second interval (or 30 to 60 frames) before the commencement of the crossing event, as specified in the dataset's annotations. For all models, the number of observation frames is 16. Sample overlap ratios are determined to be 0.6 for the PIE and JAAD datasets. Considering the training samples, which span approximately 0.5 seconds, the number of clusters is set as  \(M = 3\) and used KNN for the event merging process. All models were trained with the RMSProp optimizer with a learning rate of  $10^{-5}$, and L2 regularization with \( \lambda = 1 \times 10^{-3} \). 

The TCL model, comprising approximately 100 million parameters, was trained on the JAAD-all dataset using a cluster of eight NVIDIA 1080 Ti GPUs, setting the batch size to 2. Each training epoch took approximately 12 minutes, while inference on the JAAD-all validation set required around 2 minutes per epoch.

\subsection{Comparison with the state-of-the-art methods}
The experimental results for JAAD-beh,  JAAD-all, and PIE are shown in Tab. \ref{table:II}, Tab. \ref{table:III}, and Tab. \ref{table:IV}, respectively. We evaluated TCL’s performance by contrasting it with some other benchmark models for PCI, where different models are utilized as their encoder.

% As shown in the Tab. \ref{table:II}, where VGG and C3D indicate the use of VGG \cite{simonyan2014very} and C3D \cite{tran2015learning} as visual encoders, respectively. "GRU" and "LSTM" denote their use for processing temporal sequences. 

As shown in Tab. \ref{table:II}, TCL demonstrates exceptional performance on the JAAD-beh dataset, achieving a very high recall value of 0.98 and the best overall accuracy (ACC) of 0.74 when compared with other baseline models. It is worth highlighting that higher recall values are crucial for ensuring safety in autonomous driving. Given that JAAD-beh is particularly focused on pedestrians with the intent of crossing the road, where they exhibit more significant behavioral changes, with the TMM, the division of pedestrian postures and contextual information into distinct events becomes more manageable, enabling TCL to capture these variations more effectively. In this way, TCL can focus on crucial information among a myriad of redundant data, enabling accurate predictions of pedestrian intentions. Similar results can be found in Tab. \ref{table:III}. On the JAAD-all dataset, TCL achieves top-tier performance with an AUC of 0.92 and competitive accuracy. Moreover, it attains the highest F1 score of 0.71, reflecting a balanced trade-off between Precision and Recall—both at 0.68 and 0.74, respectively. This is primarily due to TCL's increased focus on critical events, underscoring the essential role of TMM in TCL.

% \textbf{It mainly because xxx, which further demonstrate the crucial role of xxx in xxx} 

As shown in Tab. \ref{table:IV}, TCL outperforms the most competing models on the PIE dataset, yielding the highest AUC (0.88), F1 score (0.92), and Recall (0.96). The main reason is that, by merging key events, TCL can capture critical information without causing attention dispersion. These results indicate the effectiveness of TCL in identifying PCI accurately while minimizing false negatives, which is crucial for safety in autonomous driving applications. 
%Although TCL has a slight decrease in Precision compared to the other models, which does affect its robustness and reliability.
% utilizes an impractically longer observation time of up to 1 second \cite{yao2021coupling},
% our TCL has a significant improvement in F1 score and Recall highlighting 

\begin{table}[t]
    \setlength{\abovecaptionskip}{5pt}
    \caption{The experiment results on the JAAD-beh dataset. Each metric’s top
score is emphasized in boldface, while the runner-up results are underlined.}\label{table:II}
    \centering
    \renewcommand{\arraystretch}{1.3} % 这会使行高为默认的1.25倍
    \scriptsize % Set the font size to scriptsize for the entire table
    % \begin{tabular}{p{0.15\textwidth}|p{0.08\textwidth}|p{0.02\textwidth}p{0.02\textwidth}p{0.02\textwidth}p{0.02\textwidth}p{0.01\textwidth}}
    \tabcolsep=4pt
    \begin{tabular}{l|l|l|l|l|l|l}
    \hline		
    Methods  &Encoder&AUC&ACC&F1&Prec&Rec \\  
    \hline
    % MultiRNN(CVPR18)\cite{bhattacharyya2018long} & VGG + LSTM & \\
    MultiRNN(2018) \cite{bhattacharyya2018long} & VGG + LSTM & 0.50 & -- & 0.74 & -- & -- \\
    SingleRNN(2020) \cite{kotseruba2020they} & VGG + GRU & 0.52 & 0.59 & 0.71 & 0.64 & 0.80\\
    PCPA(2021) \cite{kotseruba2021benchmark} & C3D + GRU & 0.50 & 0.58 & 0.71 & -- & --\\
    CAPformer(2021) \cite{lorenzo2021capformer} & Transformer &\underline{0.55} & -- & 0.74 & -- & -- \\
    Predicting(2022) \cite{yang2022predicting} & VGG + GRU & 0.54 & 0.62 & 0.74 & \underline{0.65} & \underline{0.85}\\
    MCIP(2022) \cite{ham2022mcip} & VGG + GRU & \underline{0.55} & 0.64 & 0.78 & -- & --\\
    Hybrid-Group(2024) \cite{dong2024pedestrian} & VGG + GRU & \bf0.61 & \underline{0.67} & \underline{0.79} & -- & --\\
    \hline
    TCL (Ours) & Transformer & \bf0.61 & \bf0.74 & \bf0.85 & \bf0.75 & \bf0.98\\
    \hline
    \end{tabular}
\end{table}

\begin{table}[t]
    \setlength{\abovecaptionskip}{5pt}
    \caption{The experiment results on the JAAD-all dataset. Each metric’s top
score is emphasized in boldface, while the runner-up results are underlined. }\label{table:III}
    \centering
    \renewcommand{\arraystretch}{1.3} % 这会使行高为默认的1.25倍
    \scriptsize % Set the font size to scriptsize for the entire table
    % \begin{tabular}{p{0.15\textwidth}|p{0.08\textwidth}|p{0.02\textwidth}p{0.02\textwidth}p{0.02\textwidth}p{0.02\textwidth}p{0.01\textwidth}}
    \tabcolsep=4pt
    \begin{tabular}{l|l|l|l|l|l|l}
    \hline		
    Methods  &Encoder&AUC&ACC&F1&Prec&Rec \\  
    \hline
    % MultiRNN(CVPR18)\cite{bhattacharyya2018long} & VGG + LSTM & \\
    MultiRNN(2018) \cite{bhattacharyya2018long} & VGG + LSTM & 0.79 & -- & 0.58 & -- & -- \\
    SingleRNN(2020) \cite{kotseruba2020they} & VGG + GRU & 0.76 & 0.79 & 0.54 & 0.44 & 0.71\\
    PCPA(2021) \cite{kotseruba2021benchmark} & C3D + GRU & 0.86 & 0.85 & 0.68 & -- & --\\
    Coupling(2021) \cite{yao2021coupling} & VGG + GRU & \bf0.92 & \underline{0.87} & \underline{0.70} & 0.66 & 0.74\\
    CAPformer(2021) \cite{lorenzo2021capformer} & Transformer & 0.70 & -- & 0.51 & -- & -- \\
    Predicting(2022) \cite{yang2022predicting} & VGG + GRU & 0.82 & 0.83 & 0.63 & 0.51 & \bf0.81\\
    MCIP(2022) \cite{ham2022mcip} & VGG + GRU & 0.84 & \bf0.88 & 0.66 & -- & --\\
    PIT(2023) \cite{zhou2023pit} & Transformer & \underline{0.89} & \underline{0.87} & 0.67 & 0.58 & \underline{0.80}\\
    TREP(2023) \cite{zhang2023trep} & Transformer & 0.86 & \bf0.88 & 0.61 & \bf0.70 & 0.54\\
    \hline
    TCL (Ours) & Transformer & \bf0.92 & \underline{0.87} & \bf0.71 & \underline{0.68} & 0.74\\
    \hline
    \end{tabular}
\end{table}

\begin{table}[t]
    \setlength{\abovecaptionskip}{5pt}
    \caption{The experiment results on the PIE dataset. Each metric’s top
score is emphasized in boldface, while the runner-up results are underlined.}\label{table:IV}
    \centering
    \renewcommand{\arraystretch}{1.3} % 这会使行高为默认的1.25倍
    \scriptsize % Set the font size to scriptsize for the entire table
    % \begin{tabular}{p{0.15\textwidth}|p{0.08\textwidth}|p{0.02\textwidth}p{0.02\textwidth}p{0.02\textwidth}p{0.02\textwidth}p{0.015\textwidth}}
    \tabcolsep=4pt
    \begin{tabular}{l|l|l|l|l|l|l}
    \hline		
    Methods  &Encoder&AUC&ACC&F1&Prec&Rec \\  
    \hline
    % MultiRNN(CVPR18)\cite{bhattacharyya2018long} & VGG + LSTM & \\
    MultiRNN(2018) \cite{bhattacharyya2018long} & VGG + LSTM & 0.80 & 0.83 & 0.71 & 0.69 & 0.73 \\
    SingleRNN(2020) \cite{kotseruba2020they} & VGG + GRU & 0.64 & 0.76 & 0.45 & 0.63 & 0.36\\
    SF-GRU(2020) \cite{rasouli2020pedestrian} & VGG + GRU & 0.83 & 0.84 & 0.72 & 0.66 & 0.80\\
    PCPA(2021) \cite{kotseruba2021benchmark} & C3D + GRU & 0.86 & 0.87 & 0.77 & -- & --\\
    Coupling(2021) \cite{yao2021coupling} & VGG + GRU & \bf0.88 & 0.84 & \underline{0.90} & \bf0.96 & \underline{0.84}\\
    CAPformer(2021) \cite{lorenzo2021capformer} & Transformer & 0.85 & -- & 0.78 & -- & -- \\
    MCIP(2022) \cite{ham2022mcip} & VGG + GRU & \underline{0.87} & \underline{0.89} & 0.81 & -- & --\\
    PFRN(2024) \cite{lv2024pedestrian} & VGG + GRU & 0.85 & \bf0.90 & 0.77 & 0.81 & 0.74\\
    \hline
    TCL(Ours) & Transformer & \bf0.88 & 0.87 & \bf0.92 & \underline{0.89} & \bf0.96\\
    \hline
    \end{tabular}
\end{table}

% Ablation studies on the proposed model using the PIE dataset assess the impact of various components, including non-visual (NV) input features, TMM, and CAB.

\begin{table}[t]
    \centering
    \setlength{\abovecaptionskip}{5pt}
    \caption{Ablation studies on the proposed model using the PIE dataset evaluate the impact of non-visual input features, TMM, and CAB. Each metric’s top
score is emphasized in boldface, while the runner-up results are underlined.}\label{table:V}
    \renewcommand{\arraystretch}{1.3}
    \scriptsize
    \tabcolsep=4pt
    \begin{tabular}{ccc|ccccccccccc}
    \hline
    NV & TMM & CAB & AUC & ACC & F1 & Precision & Recall \\
    \hline
     & \checkmark & \checkmark & 0.80 & 0.84 & 0.90 & \underline{0.88} & 0.93 \\
     \checkmark &  & \checkmark & \underline{0.86} & \underline{0.85} & \underline{0.91} & \underline{0.88} & 0.93 \\
     \checkmark & \checkmark & & 0.84 & \underline{0.85} & \underline{0.91} & 0.86 & \bf0.97 \\
    \checkmark & \checkmark & \checkmark & \bf0.88 & \bf0.87 & \bf0.92 & \bf0.89 & \underline{0.96} \\
    
    \hline
    \end{tabular}
\end{table}
\subsection{Ablation Study}
% \textbf{Effectiveness of the Temporal Merging Module: }In the ablation study presented in TABLE \ref{table:II}, we can observe the impact of incorporating the Temporal Merging module into our transformer-based model. The inclusion of Temporal Merging (Ours) leads to a noticeable enhancement in all metrics.

% \textbf{Impact of Input Features: }
In this section, we analyze the efficiency of the essential components, including non-visual input features, TMM, and CAB within TCL on the PIE dataset. Specifically, we sequentially disabled each module, while maintaining the other two modules active, resulting in three reduced TCL configurations (\ie without non-visual input features, without TMM, and without CAB). This allows the isolation and evaluation of each module’s individual contribution to the overall system performance. We report the performance of these three TCL variants in comparison with the default TCL model in Tab. \ref{table:V}.

It is evident that the inclusion of non-visual input features significantly enhances detection performance across all metrics, as demonstrated by the substantially improved performance of the TCL model and its variants that leverage these features compared to the variants without them. The integration of non-visual input features into the TMM and CAB modules results in a notable improvement in the overall performance of the default TCL framework compared to its two variants, which also include non-visual inputs. This confirms that dynamic environmental changes have a significant impact on pedestrians' motion. 

% To demonstrate the individual impact of the non-visual input features and our proposed modules, we conduct an ablation study on the proposed algorithm using the PIE dataset, systematically omitting one issue at a time. It can be observed in Tab. \ref{table:V} that the non-visual features significantly influence the accuracy of the model. The inclusion of the TMM and CAB all lead to a noticeable enhancement in most metrics. With the TMM, TCL can more accurately capture the dynamic changes in pedestrian postures. 

Additionally, in the CAB, we initially allocate attention to different events and then adaptively aggregate the event feature along with visual and non-visual data. This allows TCL to further focus on the key features that are most crucial for aligning intentions during these dynamic changes. Besides, in the TMM, event clustering ensures that the distinctions between events are significantly greater than those between individual frames. The pronounced variation in pedestrian posture and environmental information enables TCL to avoid attention dispersion due to data redundancy, allowing it to focus more effectively on the dynamic changes related to pedestrians'
action. Using the TMM alone achieves the highest recall at 0.97, showcasing TMM's proficiency in detecting samples where pedestrians intend to cross.

% In contrast, integrating the TMM with the CAB yields a harmonized balance across all metrics, resulting in the most advantageous overall performance, as evidenced by an AUC of 0.88, an accuracy of 0.87, an F1 score of 0.92, and a precision of 0.89.
\begin{figure}[t]
	\centering
	\includegraphics[width=0.6\textwidth]{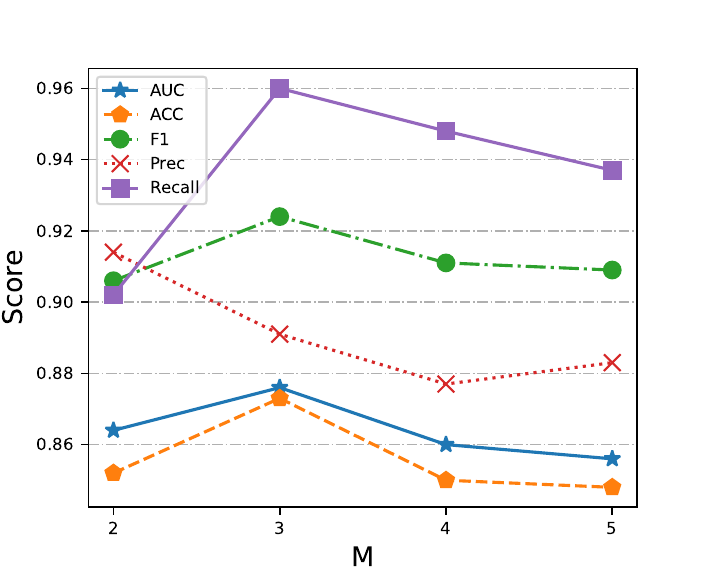}
	\caption{Performance of TCL w.r.t the cluster number \(M\) in TMM.  }\label{fig:3}
\end{figure}

\subsection{Effectiveness of the Cluster Number}
% As illustrated in Fig. \ref{fig:3}, the performance of TCL is evaluated across several metrics by varying the cluster number $M$ in TMM. 
To analyze the impact of the number of clusters $M$ on TCL's performance, we evaluate TCL's performance by varying $M$ from 2 to 5 and report the results in Fig. \ref{fig:3} across all the metrics above. 
For all metrics except for precision, TCL's performance initially displays an increasing trend, achieving peak performance when $M$ is set to 3, and then declines. Although TCL yields higher precision when $M$ is set to 2, the recall is significantly lower than when $M$ is set to 3, resulting in less optimal overall performance. Therefore, for this task, setting $M$ to 3 enables the TMM to segment events most effectively, leading to the best overall performance. On the other hand, setting $M$ higher than necessary may also hinder the effectiveness of TCL, as indicated by the declining trend. This is because, within a relatively short time frame, increasing the number of events diminishes their distinctiveness, making critical event identification less meaningful and resulting in a sharp decline in performance.

\section{Conclusion}

In this paper, we introduce a new PCI method named TCL, which synthesizes temporal feature extraction and contextual attention to merge both visual and non-visual temporal features for PCI. The proposed TCL is implemented by the TMM and the CAB. The TMM merges temporal features into limited key events, and the CAB employs the attention mechanism to fuse contextual features at both the event level and the data level. This allows the TCL to focus on critical dynamic changes, which is important for PCI. 
Extensive experiments demonstrate the superiority of our TCL compared to state-of-the-art approaches across several datasets. Specifically, the proposed TCL achieves the highest recall values, i.e., 0.96 and 0.98 on the PIE and JAAD-beh datasets, respectively. Overall, this research establishes a new benchmark in pedestrian crossing prediction and contributes to the advancement of autonomous driving systems, promoting safer and more reliable navigation.

While the TCL model holds promise for future applications like vehicle behavior prediction, its performance is contingent upon the quality and diversity of the training data. This can be a significant obstacle in real-world settings. To address this, future research should explore expanding the model's robustness by considering incorporating pedestrian trajectory information and domain adaptation capabilities and by testing more datasets under different environmental conditions.

\begin{credits}
\subsubsection{\ackname} This work is supported in part by the Chongqing Key Project of Technological Innovation and Application (No. CSTB2023TIAD-STX0015, CSTB2023TIAD-STX0031), and by the Key Cooperation Project of Chongqing Municipal Education Commission (HZ2021008).

\end{credits}

\bibliographystyle{splncs04}
\bibliography{5110}

\end{document}